\documentclass[10pt,twocolumn,letterpaper]{article}

\usepackage{cvpr}              

\usepackage[accsupp]{axessibility} 

\definecolor{cvprblue}{rgb}{0.21,0.49,0.74}
\usepackage[pagebackref,breaklinks,colorlinks,allcolors=cvprblue]{hyperref}
\usepackage{siunitx}
\usepackage{bm}
\usepackage{graphicx}
\usepackage{makecell}

\def\paperID{14782} 
\def\confName{CVPR}
\def\confYear{2026}

\title{Breaking the Scalability Limit of Multi-Projector Calibration\\ with Embedded Cameras}

\author{Takumi Kawano \quad Kohei Miura \quad Daisuke Iwai\\
The University of Osaka\\
{\tt\small https://www.xr.sys.es.osaka-u.ac.jp/}
}

\begin{document}
\maketitle
\begin{abstract}
Conventional multi-projector calibration requires projecting and capturing structured light patterns for each projector sequentially, causing calibration time and effort to increase linearly with the number of projectors. This scalability bottleneck has long limited the deployment of large-scale projection mapping systems. We present a new calibration framework that breaks this limitation by embedding cameras into the surface of the calibration target. The embedded cameras directly capture the incoming projection light, enabling the separation of simultaneously projected structured light patterns from multiple projectors according to their incident directions. Our method establishes correspondences between the optical centers of the embedded cameras and the projector pixels, allowing the intrinsic and extrinsic parameters of all projectors to be simultaneously estimated. We further introduce a correction technique for small misalignments between the calibration board and camera optical centers. As a result, our system achieves calibration accuracy comparable to conventional methods while reducing the required number of projection-capture cycles from linear to nearly constant with respect to the number of projectors, dramatically improving scalability for dense multi-projector systems with overlapping projection regions, such as high-brightness stacking, super-resolution, light-field, and shadow-suppression displays.
\end{abstract}    
\section{Introduction}
\label{sec:intro}

Projection mapping (PM)~\cite{bimber2005spatial} provides an extended reality (XR) experience by overlaying synthetic content onto physical surfaces.
Modern PM applications increasingly rely on systems composed of multiple projectors---sometimes dozens---to achieve sufficient coverage, brightness, and resolution on large-scale architectural surfaces~\cite{mine2012projection}, immersive entertainment environments~\cite{jones2014roomalive}, and complex prototypes in industrial design~\cite{cascini2020industry,yoshida2025tvcg,takezawa2019ieeevr}.
Extreme cases include light-field display systems with hundreds of co-located projectors~\cite{lee2013optimal}, as well as high-brightness stacking and super-resolution displays where all projectors share a common projection region.
To accurately align the projected imagery with the target surface, geometric calibration of each projector is essential.
Since a projector can be modeled as an inverse camera, its calibration can, in principle, follow similar methods used for camera calibration.
However, unlike cameras, projectors cannot directly observe the 3D locations of pixels projected in a physical scene, which necessitates a different calibration strategy.

In practice, projector calibration is performed indirectly by capturing structured light patterns projected onto a calibration target using an external camera.
While this process works well for a single projector, it becomes considerably more challenging when multiple projectors are involved.
When several projectors simultaneously project structured light patterns, their images can overlap on the target surface, making the individual patterns indistinguishable in the captured images.
Consequently, conventional approaches must calibrate projectors sequentially, one at a time, to avoid pattern interference.
This causes the calibration time and effort to increase linearly with the number of projectors---a fundamental scalability bottleneck that severely hinders the practical deployment of large-scale projector arrays.
Therefore, reducing this calibration overhead has become as critical as achieving high accuracy.

\begin{figure}[t]
    \centering
    \includegraphics[width=\linewidth]{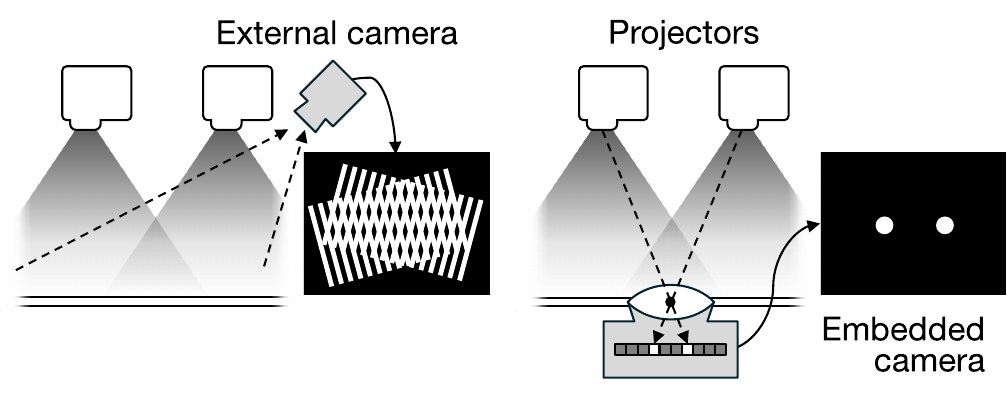}
    \makebox[0.49\hsize][c]{\raisebox{0.5ex}[0ex][0ex]{\small (a) Conventional technique}}
    \makebox[0.49\hsize][c]{\raisebox{0.5ex}[0ex][0ex]{\small (b) Proposed technique}}
    \caption{Imaging principle of simultaneously projected structured light patterns from multiple projectors: (a) Overlapping patterns captured by an external camera are difficult to separate, whereas (b) the embedded cameras directly receive the projection light, enabling separation of individual patterns.}
    \label{fig:concept}
\end{figure}

To address this challenge, we propose a parallel calibration method that allows all projectors to simultaneously project structured light patterns that can still be separated and decoded independently (Fig.~\ref{fig:concept}).
In conventional setups, patterns from different projectors cannot be separated because the calibration surface diffusely reflects the incident light, causing the directional information of each ray to be lost.
To overcome this limitation, we embed cameras so that their optical centers are aligned with the calibration plane, allowing them to directly capture light from each projector.
Since the incident direction determines the position of the light on the camera's image sensor, the system preserves directionality even when multiple patterns overlap on the surface, enabling per-projector decoding from a single capture.

While effective, this configuration also introduces a unique technical issue: the optical centers of the embedded cameras are slightly offset from the calibration plane in practice.
We propose an algorithm that compensates for this issue and validate the system through hardware implementation, showing that it achieves accuracy comparable to existing methods while reducing calibration time to nearly constant with respect to the number of projectors, provided that all projectors share a common projection region.

\section{Related Work}

Projector calibration techniques have been extensively studied not only in the context of PM but also for 3D shape measurement.
In both applications, the underlying calibration principle is consistent: a projector is modeled as an inverse camera, and the same mathematical formulation used for camera calibration is applied.
%
%
Specifically, the well-known pinhole camera model is employed, where, given a 3D point $(X, Y, Z)$, the projected pixel $(x, y)$ is computed as $s[x\; y\; 1]^T = K[R\; |\; t][X\; Y\; Z\; 1]^T$, where $s$ is an arbitrary scale factor.
$K = [f_x\; 0\; c_x;\; 0\; f_y\; c_y;\; 0\; 0\; 1]$ is the intrinsic matrix composed of the focal lengths $(f_x, f_y)$ and optical center $(c_x, c_y)$, while the skew is typically ignored following prior works.
The extrinsic matrix $[R\ |\ t]$ represents the projector's pose in the world coordinate system.

Four major approaches have been proposed for estimating intrinsic and extrinsic parameters.  
The first approach estimates parameters by collecting multiple 3D-2D correspondences between points in the scene and their corresponding projector pixels.
The 3D coordinates of the scene can be obtained either by using a calibration target with known geometry~\cite{8466021} or by estimating them using stereo cameras~\cite{809883,1640446} or RGB-D sensors~\cite{10458348,jones2014roomalive}.
The projector's image coordinates are then determined by projecting structured light patterns (e.g., Gray-codes) or 2D markers such as ArUco markers onto the scene and capturing them with an external camera.

The second approach uses a planar calibration target printed with markers such as checkerboard patterns, similar to those used in camera calibration~\cite{888718}.
The target is placed at multiple poses to cover the projector's field of view, and 2D-2D correspondences between printed markers and projected image coordinates are collected for calibration.
As in the first approach, the projector image coordinates are obtained by projecting structured light patterns~\cite{6375029,10.1117/1.2336196,9107397} or 2D markers~\cite{4270468,1565408,5204319,10.1117/1.2969118} and observing them with an external camera.
In studies that use projected 2D markers, different color combinations between printed and projected patterns have been utilized to allow stable separation of the two patterns in captured images.

The third approach is self-calibration~\cite{5981781,8115403,7164353,7815278}, which estimates the intrinsic and extrinsic parameters of a projector-camera pair without prior knowledge about the scene surface shapes. 
These methods typically project structured light patterns onto a 3D scene to obtain projector-camera correspondences, from which the fundamental matrix is estimated and then decomposed into the intrinsic and extrinsic parameters of both devices.

The fourth and most recent line of work employs differentiable frameworks to estimate calibration parameters.
These methods typically project a variety of pattern images onto the scene while capturing them from multiple viewpoints using a handheld camera.
The 3D shape of the scene is then reconstructed using techniques such as NeRF~\cite{10269043} or 3D Gaussian Splatting~\cite{11192660}, from which the calibration parameters are subsequently estimated.

All of these approaches rely on external cameras to capture spatial patterns projected onto a calibration target or the scene.
When patterns from multiple projectors overlap in the scene, it becomes difficult to separate them, making simultaneous calibration of multiple projectors infeasible.
Therefore, each projector must be calibrated sequentially by projecting and capturing patterns one at a time.
As a result, the calibration time increases linearly with the number of projectors, requiring significant effort and complex procedures for systems involving many projectors.

In this work, we address the long-standing scalability problem in multi-projector calibration---a fundamental issue that, to our knowledge, has remained unsolved for over four decades.
We overcome the inability to separate simultaneously projected patterns from multiple projectors by adopting an inverse setup, both conceptually and physically: orienting cameras toward the projectors and treating them as the calibration target rather than as sensors observing it.
Our approach draws inspiration from two lines of prior work: (1) the use of pinhole arrays to separate spatial patterns for calibrating the intrinsic parameters of a single camera~\cite{8099504} or a single projector~\cite{9523844}, and (2) the use of light sensors embedded within the scene to directly capture structured light from projectors for single-projector registration~\cite{8007248,10.1145/1029632.1029653,10.1145/1015706.1015738}.
In contrast, to the best of our knowledge, our method introduces a fundamentally new paradigm that enables the simultaneous calibration of multiple projectors, estimating both their intrinsic and extrinsic parameters.

\section{Method}

We present a method that breaks the scalability limit of multi-projector calibration by embedding cameras into the surface of the calibration target.
The embedded cameras directly capture the incoming projection light, enabling the separation of simultaneously projected structured light patterns from multiple projectors according to their incident angles.
Without loss of generality, we describe our method using a planar checkerboard, a commonly used calibration target for camera calibration, and base our formulation on the widely used method of Zhang~\cite{888718}.
Both the cameras and projectors are modeled using the pinhole camera model, which serves as the foundation of the subsequent formulation.
Note that the proposed technique is not limited to planar targets; it can be directly applied to calibration targets composed of multiple planar faces (e.g., polyhedral targets).

\subsection{Geometric Configuration and Calibration Principle}\label{subsec:calib-principle}

\begin{figure}[t]
  \centering
  \includegraphics[width=\linewidth]{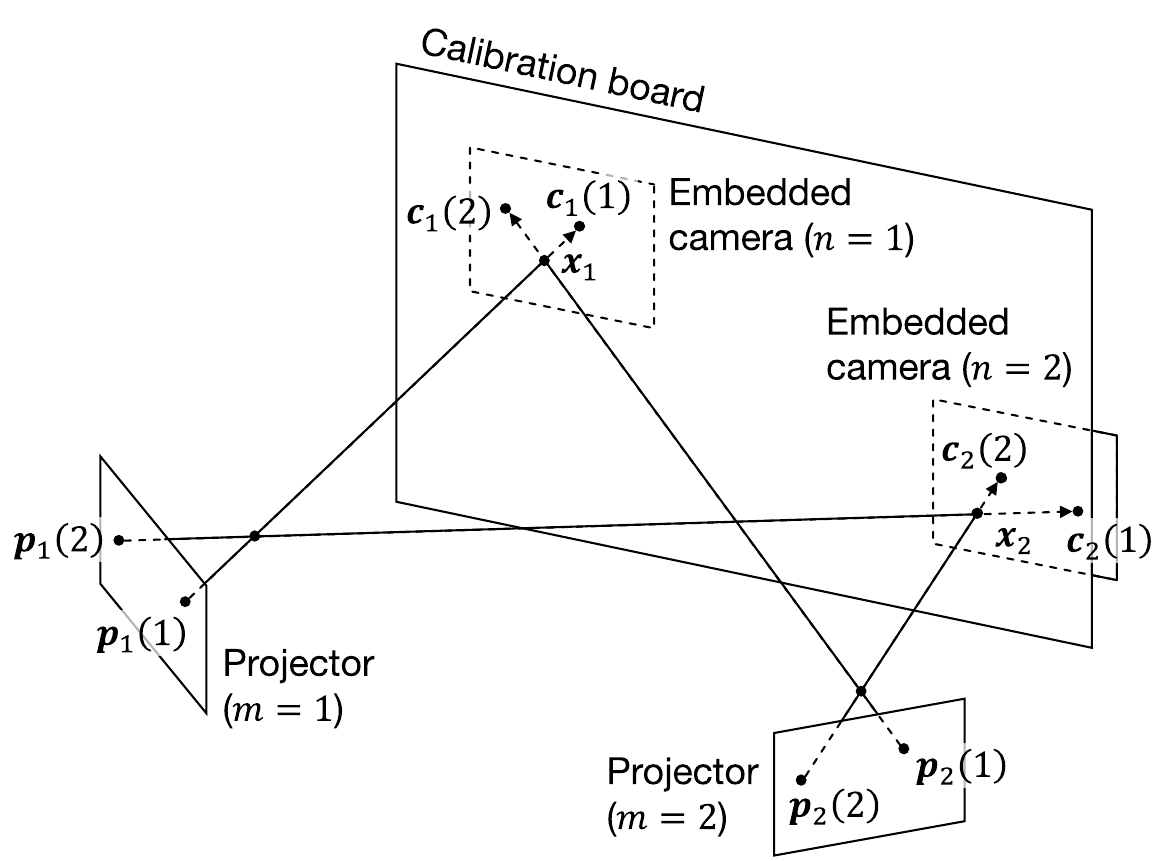}
  \caption{Geometric relationship of light rays in the proposed method for the case of two projectors ($M=2$) and two cameras ($N=2$).}
  \label{fig:principle}
\end{figure}

We replace the printed checkerboard pattern with $N$ cameras whose optical centers are aligned with the plane of the board.
When $M$ projectors simultaneously project structured light patterns onto the calibration board, the projected light from all projectors overlaps on the board surface.
Let us focus on the light from a projector $m\ (=1,\ldots,M)$ that is observed by a camera $n\ (=1,\ldots,N)$.
This light travels along the line that passes through both optical centers of the projector $m$ and the camera $n$.
The light emitted from the projector pixel $\bm{p}_m(n)$, which lies at the intersection of this line with the projector image plane, enters the camera pixel $\bm{c}_n(m)$, located at the intersection of the same line with the camera image plane (Fig.~\ref{fig:principle}).

Because the light rays from projectors located at different positions reach different pixels on the camera image plane, the incident light beams that overlap at the optical center can still be optically separated and observed independently.
This directional-encoding property is analogous to the principle of light-field cameras~\cite{ng2005light}.
This enables us to obtain the correspondence between the 2D coordinate $\bm{x}_n$ of each camera's optical center defined in the board coordinate system and its corresponding projector image coordinate $\bm{p}_m(n)$.
By capturing this correspondence from multiple board poses, the intrinsic and extrinsic parameters of each projector can be estimated using Zhang's method~\cite{888718}.

To achieve subpixel calibration accuracy, we use not only Gray-code patterns but also line-shift patterns~\cite{10.1117/12.410877} as structured light.
In the line-shift projection, only a small neighborhood around each projector pixel $\mathbf{p}_m(n)$, whose location is estimated from the Gray-code decoding is probed.
Specifically, several vertical and horizontal white lines are projected while shifting them pixel by pixel in the $x$- and $y$-directions around $\mathbf{p}_m(n)$.
In addition to these patterns, we project a binary temporal sequence of uniform-white and uniform-black images encoding the projector ID $m$, which allows us to identify which projector emitted the light observed at each camera pixel.

Each camera $n$ captures these temporal sequences and obtains $\bm{p}_m(n)$, the coordinate of a pixel in projector $m$ illuminating the camera's optical center, according to the following procedure.
First, for each camera pixel, the maximum and minimum intensity values across the captured image sequence are extracted.
If their difference exceeds a threshold $t$, the pixel is marked as receiving light from a projector.
Next, for each such pixel, the emitting projector $m$ is identified by decoding the binary temporal sequence pattern, giving $\bm{c}_n(m)$.
Finally, the temporal intensity sequences in the captured Gray-code and line-shift patterns at $\bm{c}_n(m)$ are used to decode the projector coordinate $\bm{p}_m(n)$.
In practice, due to defocus blur and optical aberrations, light from a single projector pixel of projector $m$ is potentially observed by multiple pixels of camera $n$, resulting in multiple $\bm{c}_n(m)$.
In such cases, we average the intensity values across those pixels for decoding $\bm{p}_m(n)$.

\subsection{Optical Center Estimation and Misalignment Compensation}\label{subsec:comp-misalignment}

When embedding cameras into the calibration board, ensuring that their optical centers lie precisely on the board surface is difficult: compound optics make the exact optical center position indeterminate, and even a known offset cannot practically be zeroed out.
Any deviation causes $\bm{x}_n$ to be estimated incorrectly, degrading calibration accuracy.

\begin{figure}[t]
  \centering
  \includegraphics[width=0.85\linewidth]{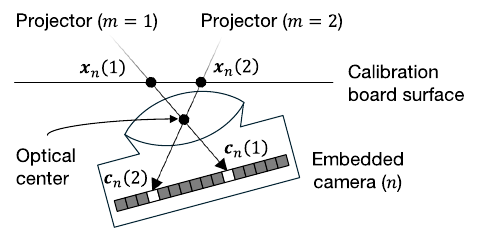}
  \caption{The light ray from projector $m$ observed by camera $n$ intersects the plane of the calibration board at $\bm{x}_n(m)$ and the camera image plane at pixel $\bm{c}_n(m)$.}
  \label{fig:comp-misalignment}
\end{figure}

When the optical center does not lie on the board surface, the light ray from projector $m$ to camera $n$ intersects the board at a different point $\bm{x}_n(m)$ for each projector (Fig.~\ref{fig:comp-misalignment}).
To address this, we assume that the optical centers are not perfectly aligned with the board surface and modify the calibration procedure described in Section \ref{subsec:calib-principle} accordingly.
Specifically, we use $\bm{x}_n(m)$ instead of $\bm{x}_n$ as the board coordinate input to Zhang's method~\cite{888718}.
We derive $\bm{x}_n(m)$ from the camera pixel $\bm{c}_n(m)$ that receives the projected light from projector $m$.
Let the mapping between these two coordinates be denoted by $\mathcal{M}_n$:
\begin{equation}
    \bm{x}_n(m) = \mathcal{M}_n(\bm{c}_n(m)).
\end{equation}
From geometric optics, $\mathcal{M}_n$ is a projective transformation, and thus, can be modeled as a homography transformation.

The parameters of $\mathcal{M}_n$ are determined through the following offline procedure.
First, a printed checkerboard pattern is attached to the calibration board to define a real-scale 2D coordinate system.
In implementation, holes are made in the board at the camera installation positions so that the projection light from the projector can pass through to the cameras.
Next, a single projector is prepared and placed at an arbitrary 3D position $\bm{X}_k$ in space.
The projector projects Gray-code and line-shift patterns onto the board.
For each embedded camera $n$, the coordinate $\bm{p}_k(n) = [u_k(n)\ v_k(n)]^T$ of the observed projector pixel is recorded, together with the corresponding camera pixel $\bm{c}_n(k)$ that captures it.
When multiple camera pixels observe the same projector pixel (due to defocus blur or aberrations), their centroid is used as $\bm{c}_n(k)$.

\begin{figure}[t]
  \centering
  \includegraphics[width=\linewidth]{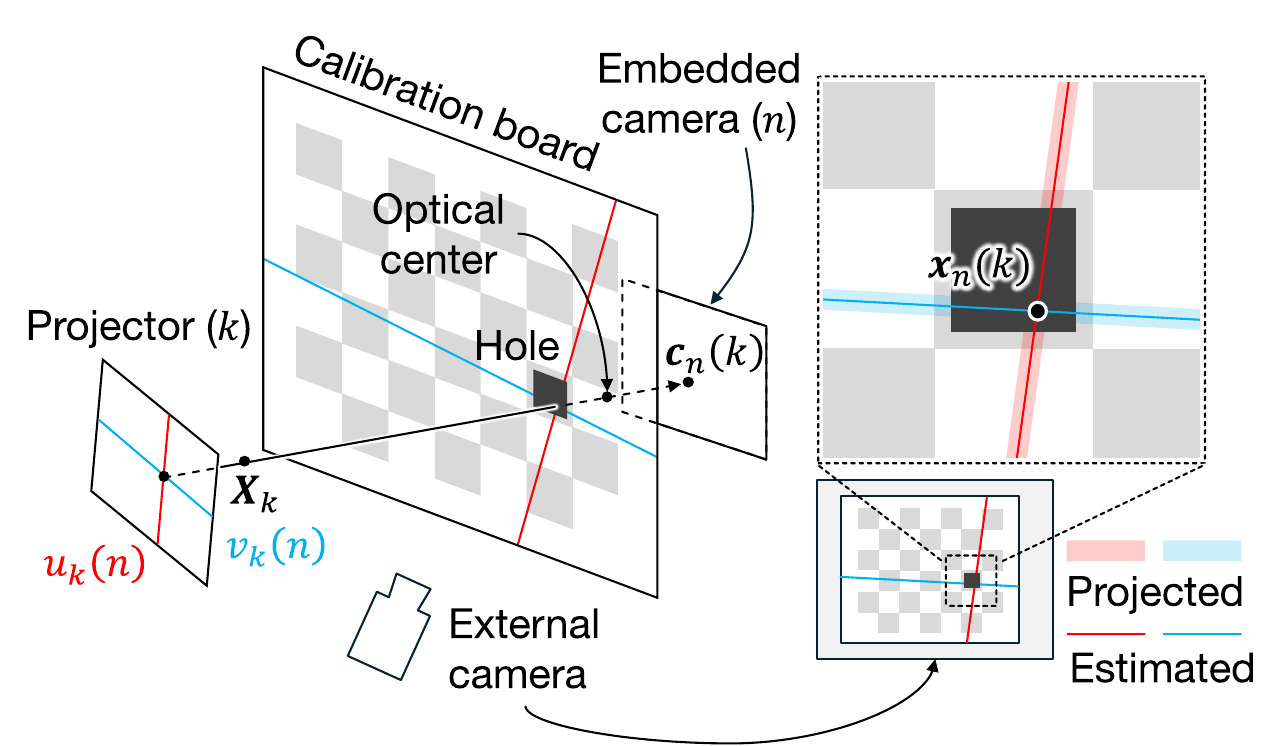}
  \caption{Measurement method of the intersection point $\bm{x}_n(k)$ between the projection light from a projector at $\bm{X}_k$ to an embedded camera $n$ and the calibration board plane. (In implementation, the red and blue lines are projected separately.)}
  \label{fig:comp-principle}
\end{figure}

Then, an image consisting of a single vertical white line at $x = u_k(n)$ on an otherwise black background is projected, and an external camera captures the projected line on the calibration board (Fig.~\ref{fig:comp-principle}).
After binarizing the captured image and detecting the line pixels, the 2D line equation of the projected line in the board coordinate system is computed using the Hough transform.
Similarly, a horizontal line at $y=v_k(n)$ is projected, and the same process is repeated to obtain the equation of the horizontal line.
The intersection of these two lines is computed and recorded as $\bm{x}_n(k)$.
By repeating this process for $K$ different projector positions, we obtain $K$ pairs of corresponding coordinates $\{\bm{c}_n(k), \bm{x}_n(k)\}$.
Finally, the homography matrix representing the transformation $\mathcal{M}_n$ is estimated using a least-squares method with RANSAC.
This offline calibration is performed only once after embedding the cameras into the calibration board.

\noindent\textbf{Lens distortion.}
Since $\mathcal{M}_n$ is estimated from camera images, camera lens distortion could in principle affect the accuracy of the homography.
In practice, however, the full camera image is mapped onto an small ($\approx$ 25\,mm$^2$) region of the board, so the effective distortion within this small image area is negligible.
Incorporating explicit lens undistortion as a preprocessing step is left as future work.

\subsection{Advantages over Conventional Methods}

The proposed method provides two major technical advantages over conventional projector calibration techniques.
The first advantage is a significant reduction in calibration time.
In conventional methods, when calibrating multiple projectors, structured light patterns must be projected sequentially for each projector.
If there are $M$ projectors and structured light patterns consist of Gray-code and line-shift patterns, the number of structured light patterns that the external camera needs to capture is given by
\begin{equation}
    M\times (\lceil \log_2 W \rceil + \lceil \log_2 H \rceil + L),
\end{equation}
where $W{\times}H$ is the projector resolution and $L$ is the number of line-shift patterns.
In contrast, the proposed method projects all projectors simultaneously, reducing the required pattern count to
\begin{equation}
    \lceil \log_2 M \rceil + \lceil \log_2 W \rceil + \lceil \log_2 H \rceil + L,
\end{equation}
where $\lceil \log_2 M \rceil$, a small constant in practice (since $M \ll W, H$), accounts for projector ID identification.
Thus, calibration time scales linearly with $M$ in conventional approaches, whereas our method requires a nearly constant number of patterns.
Furthermore, since the proposed method uses a 2D calibration board similar to those widely adopted in projector calibration~\cite{5204319,6375029}, the advantage becomes even more pronounced when the board must be repositioned multiple times for calibration under different poses.

The second technical advantage is robustness against ambient light.
While PM is typically conducted in dark environments, calibration is often more convenient under bright conditions---for instance, completing daytime calibration before a night PM event.
Conventional methods rely on external cameras, and under strong ambient illumination the projected patterns are washed out, making them difficult to detect.
In contrast, our method directly receives the projection light, maintaining high contrast regardless of ambient lighting conditions.

\paragraph{Limitations.}
The method requires all projectors to share a common projection region; for wide-area systems where projectors do not overlap (e.g., building-scale projection mapping), the board must be repositioned per group, restoring linear scaling.
Non-planar targets are supported only for polyhedral shapes via a per-face homography approach; curved surfaces are left as future work.
Ambient-light robustness has been validated with two simultaneous projectors; larger configurations remain to be tested (Section~\ref{subsec:ambient-light}).

\section{Experiments}
\label{sec:exp}

\subsection{Prototype}
\begin{figure}[tb]
    \centering
    \begin{subfigure}[b]{0.48\columnwidth}
        \centering
        \includegraphics[width=\textwidth]{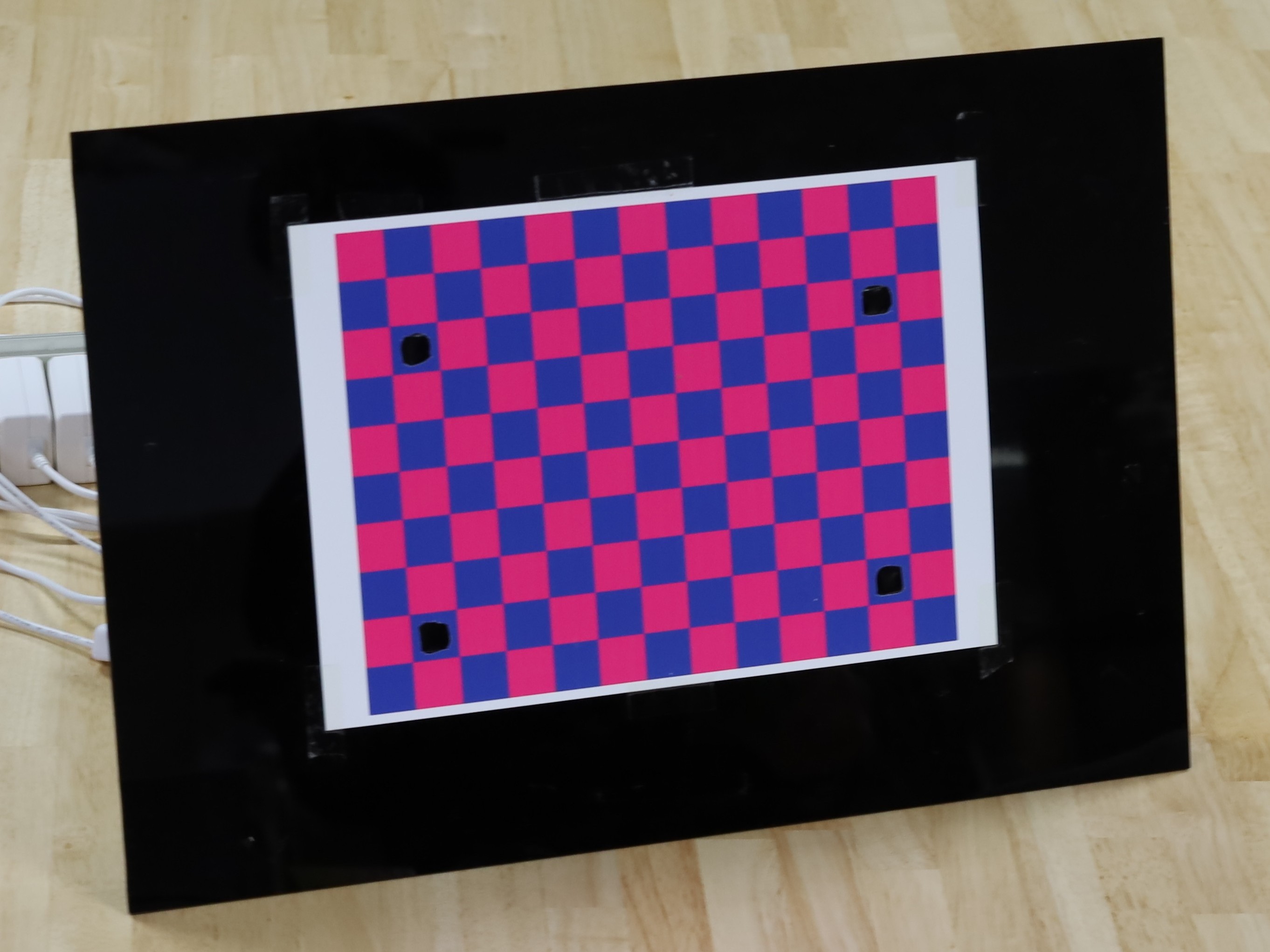}
        \caption*{Front view}
        \label{fig:prototype_front}
    \end{subfigure}
    \hfill
    \begin{subfigure}[b]{0.48\columnwidth}
        \centering
        \includegraphics[width=\textwidth]{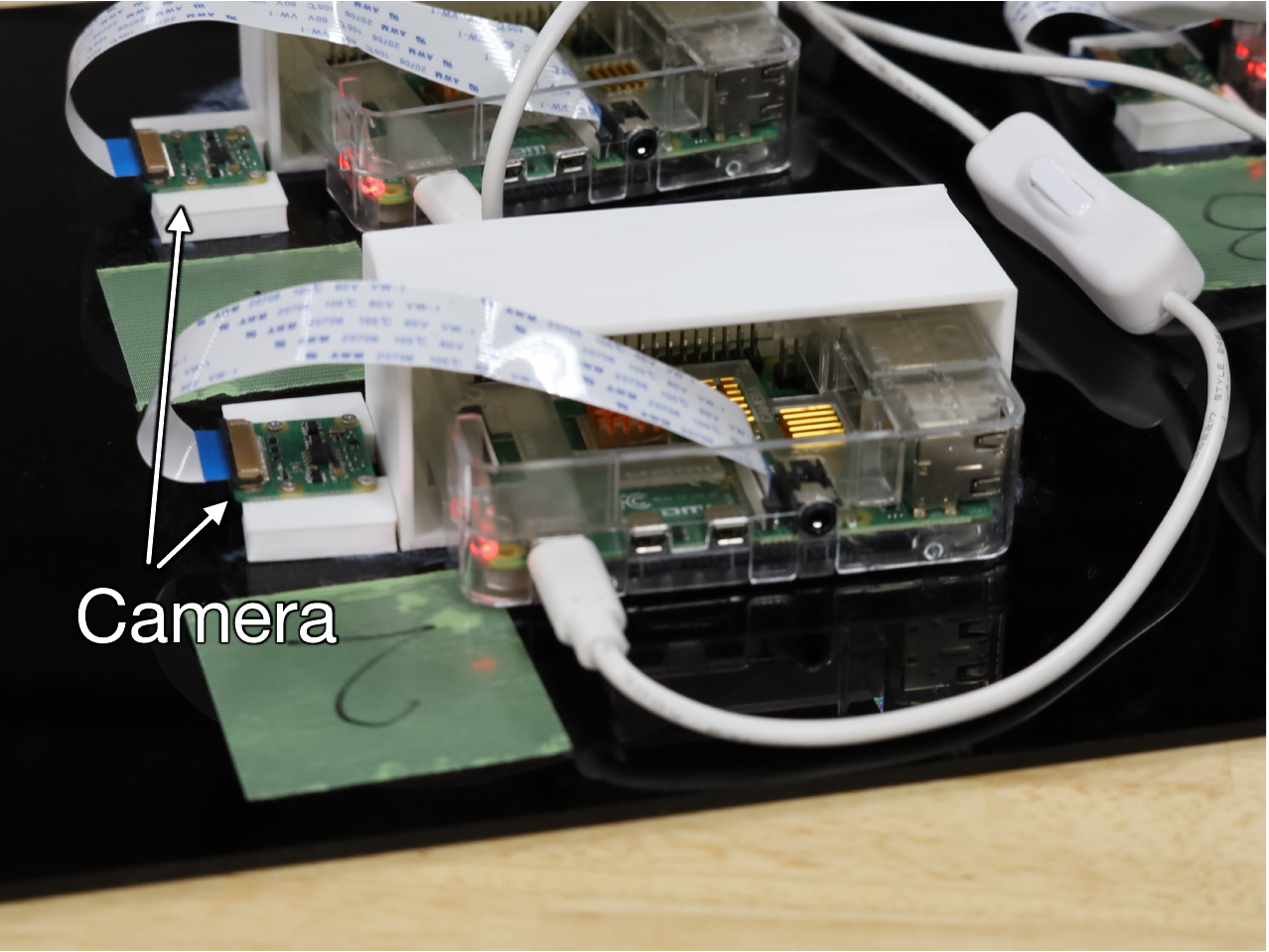}
        \caption*{Back view}
        \label{fig:prototype_back}
    \end{subfigure}
    \caption{Prototype calibration board.}
    \label{fig:prototype}
\end{figure}

We fabricated a prototype calibration board by drilling four holes into an acrylic plate measuring 470~$\times$~320~mm and inserting four wide-angle cameras (Raspberry Pi Camera Module~3 Wide, $4608 \times 2592$ pixels, $102^\circ / 67^\circ$) so that the lens tips were aligned with the board surface (Fig.~\ref{fig:prototype}). 
The four cameras were mounted at the vertices of a 200~mm~$\times$~120~mm rectangle centered on the board.
When the projector faced the board directly, the embedded cameras captured saturated bright regions around the projector lens even during black projection due to light leakage, making it difficult to distinguish between black and white projections.
To mitigate this issue, a neutral density (ND) filter (FUJIFILM ND-4.0) was placed in front of each camera lens.

To compensate for optical center misalignment of embedded cameras as described in Section~\ref{subsec:comp-misalignment}, an external camera (Canon EOS RP) was prepared, and a printed checkerboard pattern was attached to the calibration board surface. 
The checkerboard consisted of $9 \times 12$ squares, each measuring $20$ mm, alternating between blue and magenta. 
This color design allowed the checkerboard corners to be detected from the red channel of the external camera under structured light pattern projection, while the projected patterns were extracted from the blue channel. 
The holes for embedding the cameras were positioned to avoid overlapping the checkerboard corners, specifically within the blue squares. 
Next, following the procedure described in Section~\ref{subsec:comp-misalignment}, we estimated the homography transformations required to obtain the intersection points of the projection rays on the board surface, thereby compensating for optical center misalignment.
To cover the entire field of view of each embedded camera, a projector (Optoma ML1050ST+, 1280~$\times$~800 pixels) was positioned at 30--40 locations for each camera. These positions were distributed along horizontal and vertical lines across the camera's field of view, and the corresponding coordinate relationships were recorded.

\subsection{Basic Characteristics}

\begin{figure}[tb]
    \centering
    \begin{subfigure}[b]{0.48\columnwidth}
        \centering
        \includegraphics[width=\textwidth]{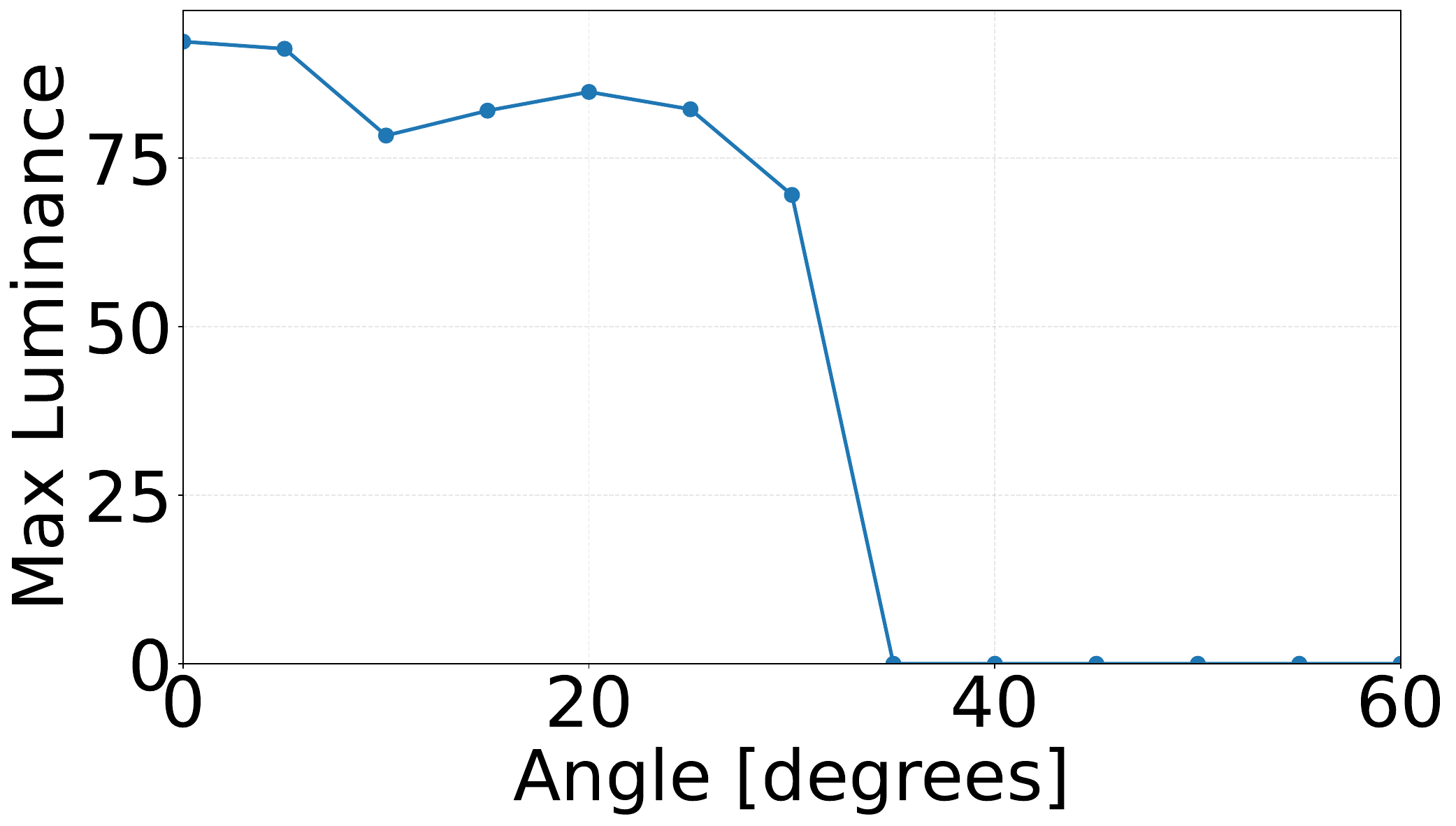}
        \caption*{\centering Rotation about the camera's $x$-axis}
        \label{fig:incident_angle_short}
    \end{subfigure}
    \hfill
    \begin{subfigure}[b]{0.48\columnwidth}
        \centering
        \includegraphics[width=\textwidth]{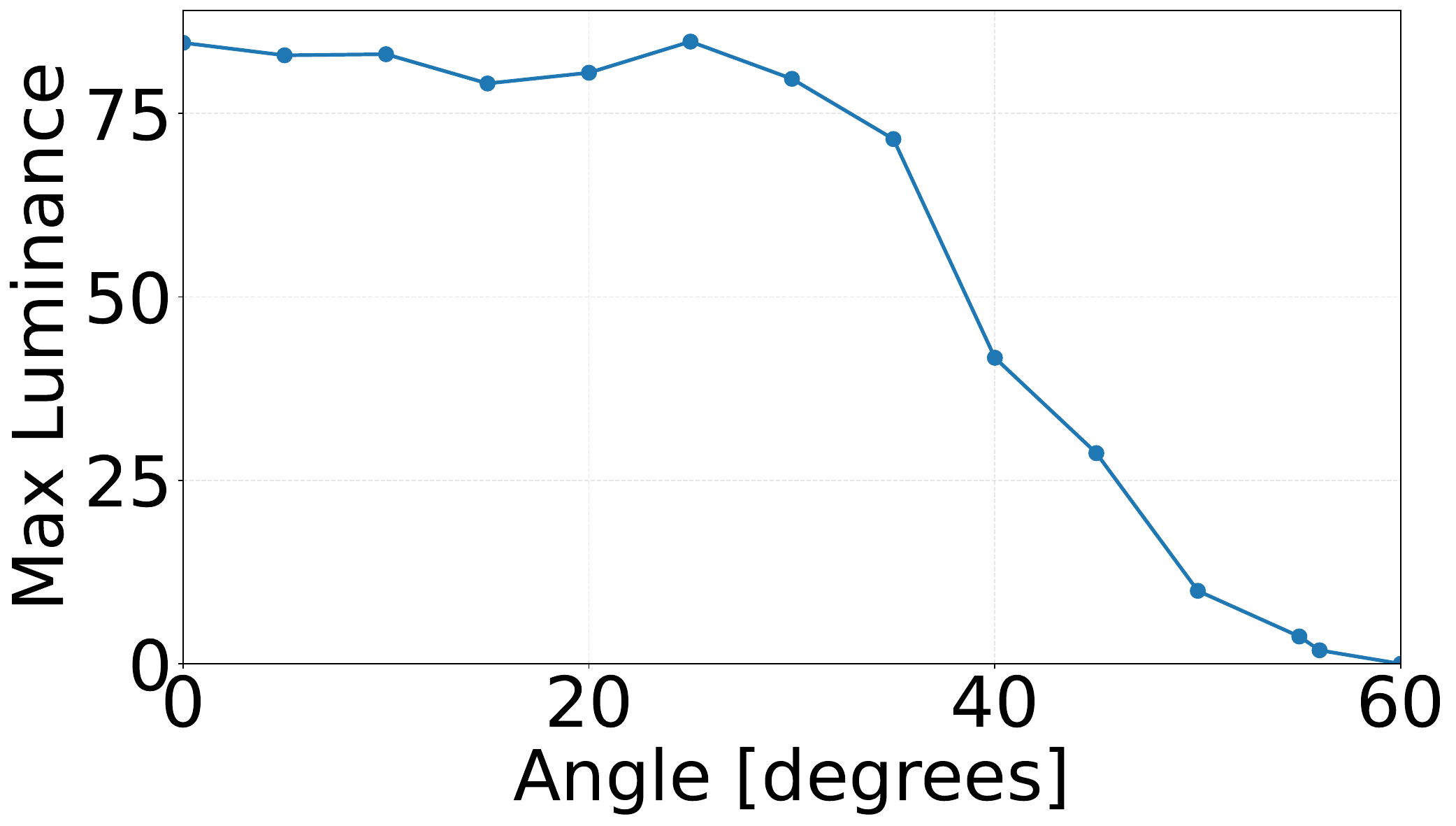}
        \caption*{\centering Rotation about the camera's $y$-axis}
        \label{fig:incident_angle_long}
    \end{subfigure}
    \caption{
        Relationship between incident angle and observed brightness by the embedded camera.
    }
    \label{fig:incident_angle}
\end{figure}

We investigated the observable range of projector positions by the embedded cameras with respect to the incident angle relative to the calibration board. Specifically, the projector was repositioned to a series of locations such that the incident angle to the board was increased in $5^\circ$ increments about the camera's $x$- and $y$-axes. 
A one-pixel-wide vertical line was scanned horizontally in the vicinity of the embedded camera, and the maximum brightness observed by the camera was recorded at each incident angle. The measured brightness for each angle is plotted in Fig.~\ref{fig:incident_angle}. 
The results show that the brightness dropped to half between $30^\circ$ and $35^\circ$ for rotation about the $x$-axis and at around $40^\circ$ for rotation about the $y$-axis, confirming that the observable incident-angle range of the embedded camera was approximately $\pm32^\circ$ about the $x$-axis and $\pm40^\circ$ about the $y$-axis.


\begin{figure}[tb]
    \centering
    \includegraphics[width=0.9\linewidth]{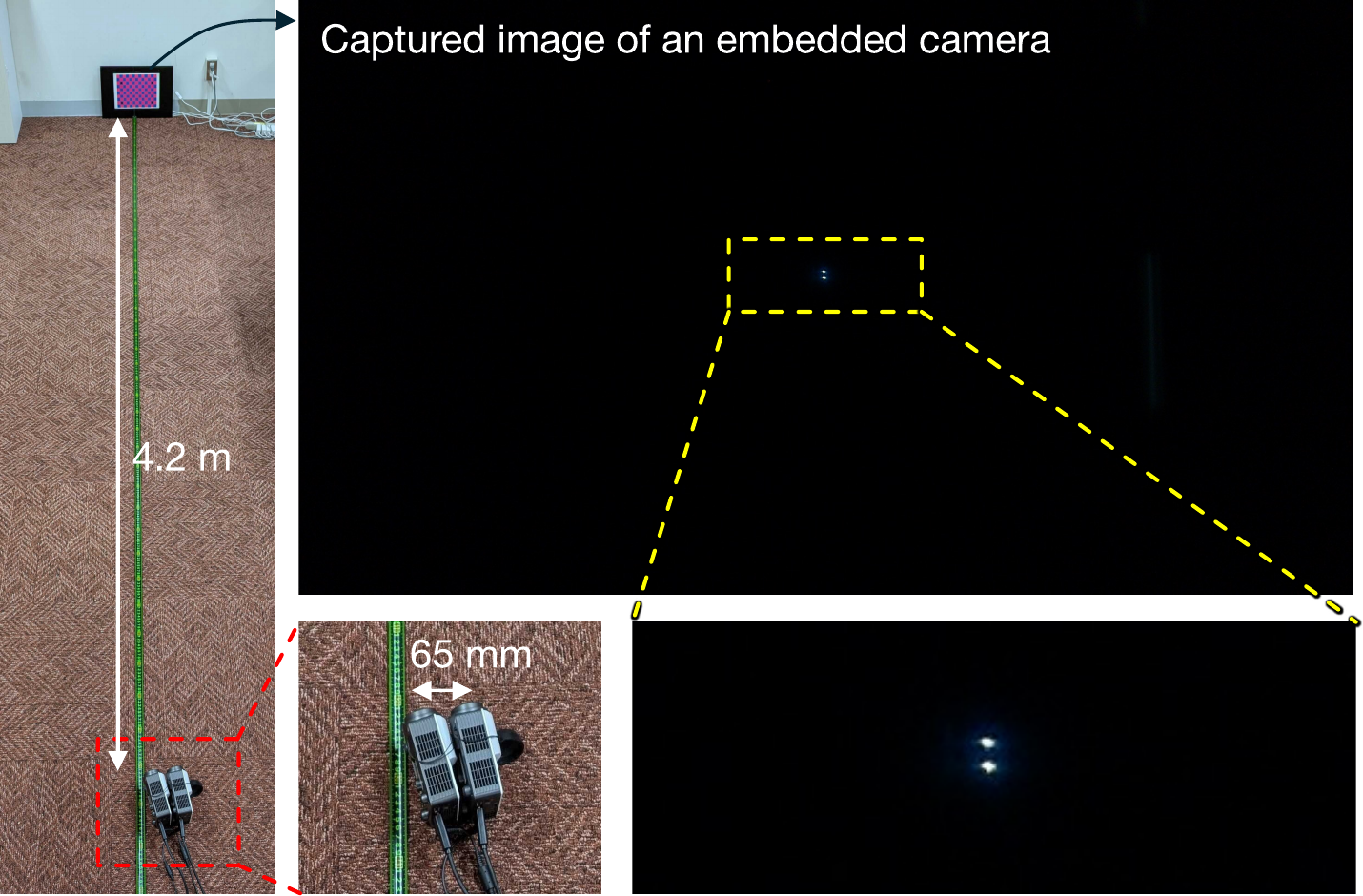}
    \caption{Experiment to determine the minimum separation distance between two projectors.}
    \label{fig:two_projector_separation}
\end{figure}

To separate light from two projectors on the embedded camera's image plane, the projectors must be placed sufficiently apart. 
We experimentally verified this by testing the limit of separability using two projectors placed as close as possible. 
Two projectors were installed 4.2~m away from the calibration board (the maximum available distance in our laboratory) and positioned such that their lenses were almost touching (65~mm apart, as measured), corresponding to an angular separation of 0.88$^\circ$ from the embedded camera's viewpoint. 
White images were projected from both projectors, and the captured image from an embedded camera is shown in Fig.~\ref{fig:two_projector_separation}. 
The two bright spots were clearly distinguishable, demonstrating that the proposed system can successfully separate and calibrate projectors even when their angular separation is below one degree.

\subsection{Multi-Projector Alignment on a Planar Surface}

We conducted two experiments to verify that the proposed method can (1) simultaneously acquire structured light patterns from multiple projectors and (2) accurately estimate the geometric correspondences between the calibration board and each projector image.

\subsubsection{Two Projectors}\label{subsubsec:2-pros}

\begin{figure}[tb]
    \centering
    \begin{subfigure}[t]{0.32\columnwidth}
        \centering
        \includegraphics[width=\textwidth]{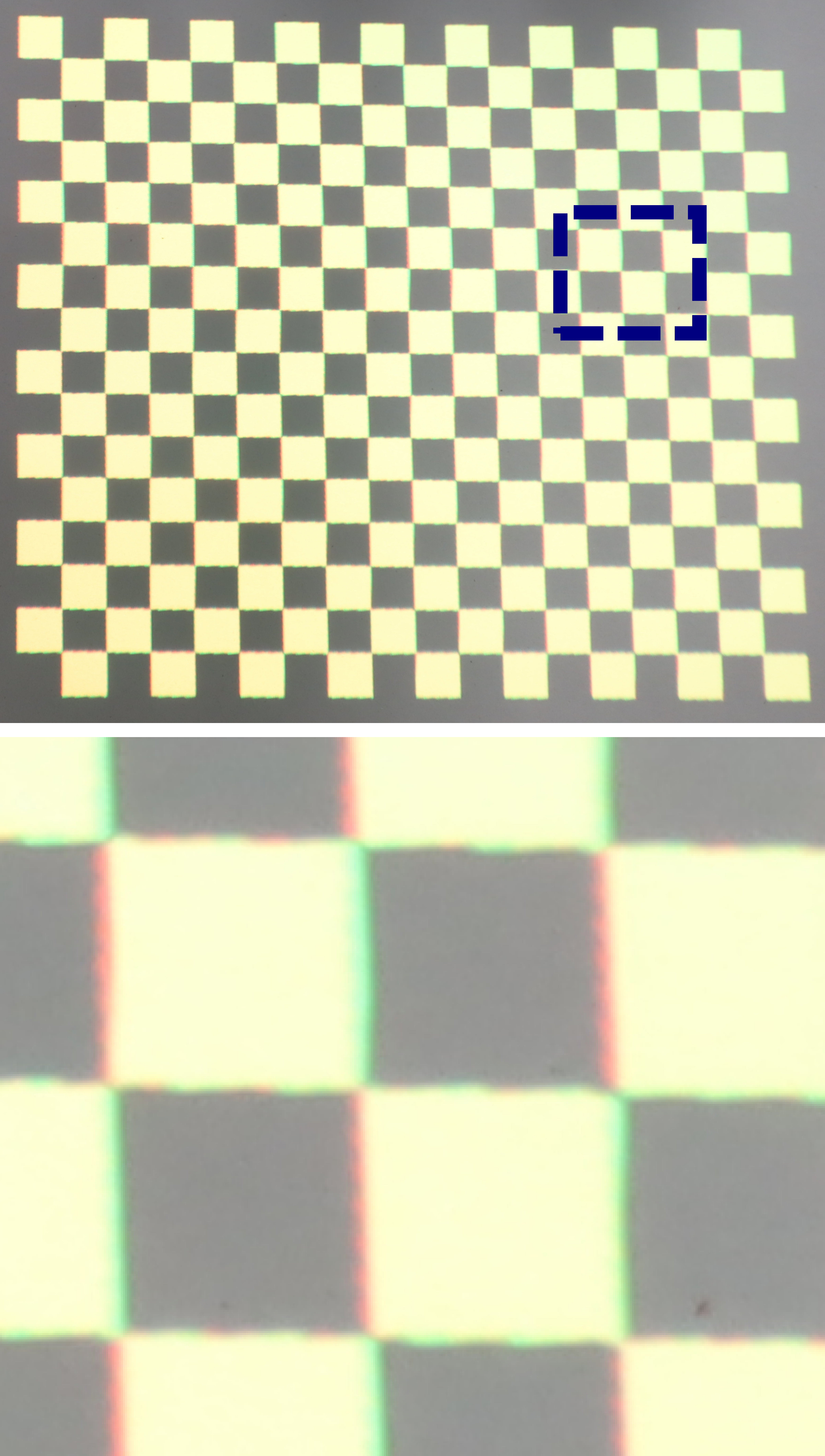}
        \caption*{\centering  Conventional}
        \label{fig:two_proj_conventional}
    \end{subfigure}
    \hfill
    \begin{subfigure}[t]{0.32\columnwidth}
        \centering
        \includegraphics[width=\textwidth]{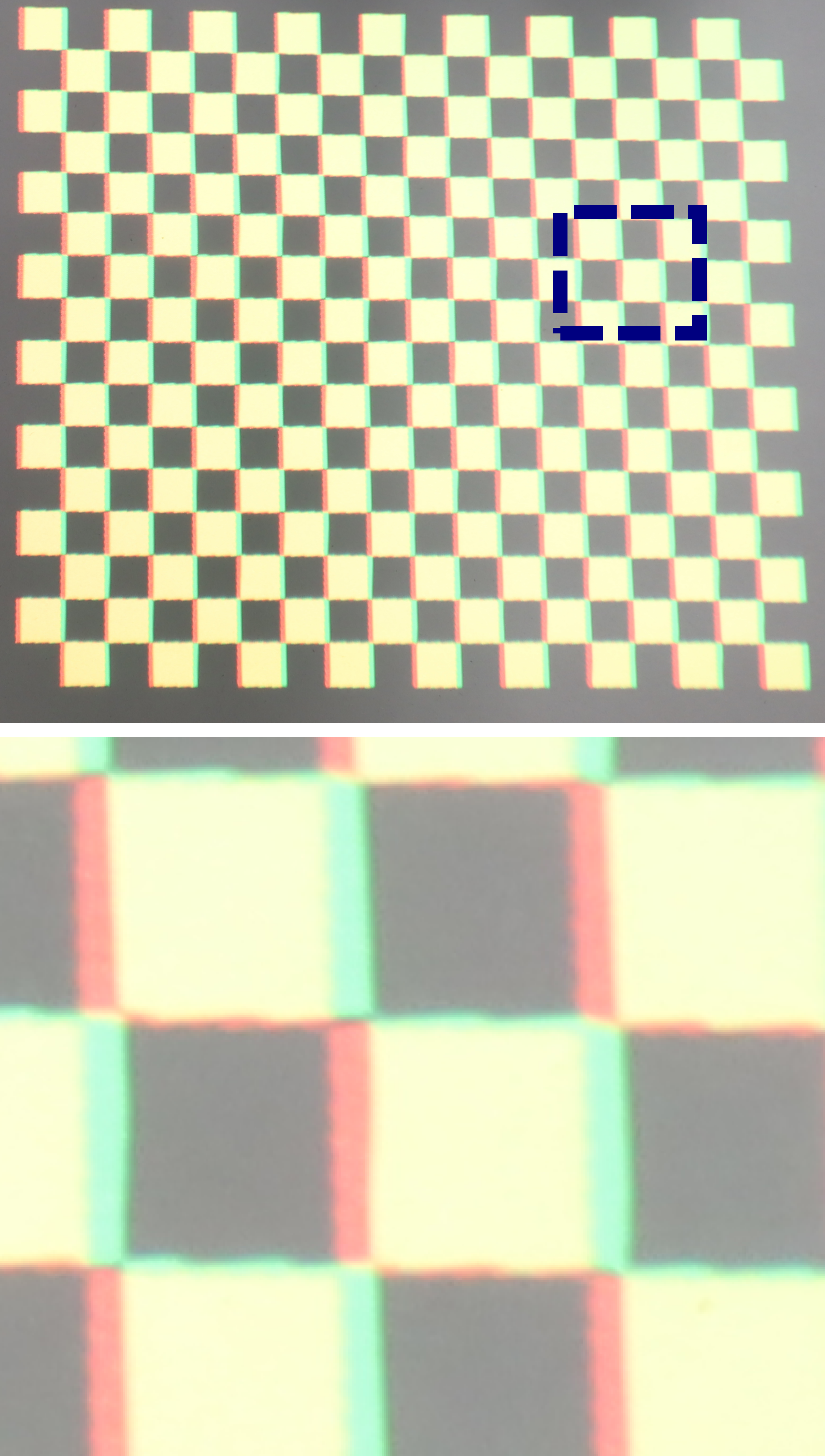}
        \caption*{ \centering Proposed\\(w/o compensation)}
        \label{fig:two_proj_uncorrected}
    \end{subfigure}
    \hfill
    \begin{subfigure}[t]{0.32\columnwidth}
        \centering
        \includegraphics[width=\textwidth]{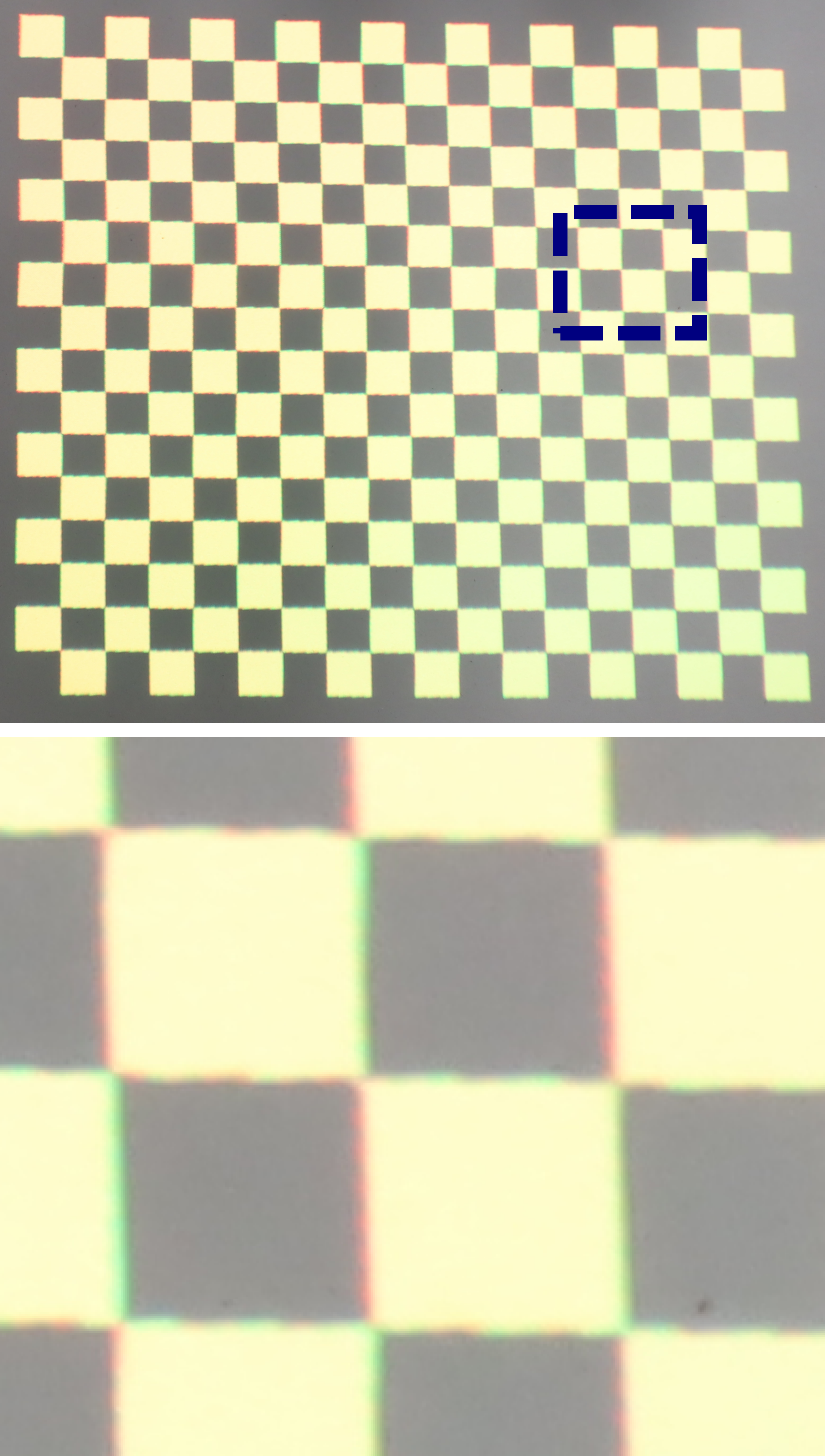}
        \caption*{\centering Proposed\\(with compensation)}
        \label{fig:two_proj_corrected}
    \end{subfigure}
    \caption{Alignment results of red and green checkerboard patterns projected from two projectors. (Top) Overall view; (Bottom) Magnified view.}
    \label{fig:two_proj_alignment}
\end{figure}

Two projectors (Optoma ML1050ST+) simultaneously projected Gray-code and line-shift patterns, which were captured by the embedded cameras. 
By capturing these images using the embedded cameras, correspondences between the projector pixel coordinates and the board coordinates were obtained. 
Homography transformations from the projector image coordinates to the calibration board coordinates were then computed under two conditions: (1) with and (2) without the compensation for optical center misalignment (Section~\ref{subsec:comp-misalignment}). 
As a baseline, we also computed homographies using a conventional method in which each projector was calibrated individually by projecting structured light patterns captured by an external camera (Canon EOS RP) (hereafter referred to as the \textit{conventional} condition).

For evaluation, a white screen was attached to the calibration board, and the two projectors projected checkerboard patterns---red from one projector and green from the other---using the computed homographies.
If the homographies were accurate, the two checkerboards should perfectly overlap, resulting in a yellow checkerboard pattern. 
As shown in Fig.~\ref{fig:two_proj_alignment}, without the compensation for optical center misalignment, the two checkerboards were misaligned, while with compensation, they overlapped precisely, achieving comparable accuracy to the conventional condition.

\begin{figure}[tb]
    \centering
    \includegraphics[width=\linewidth]{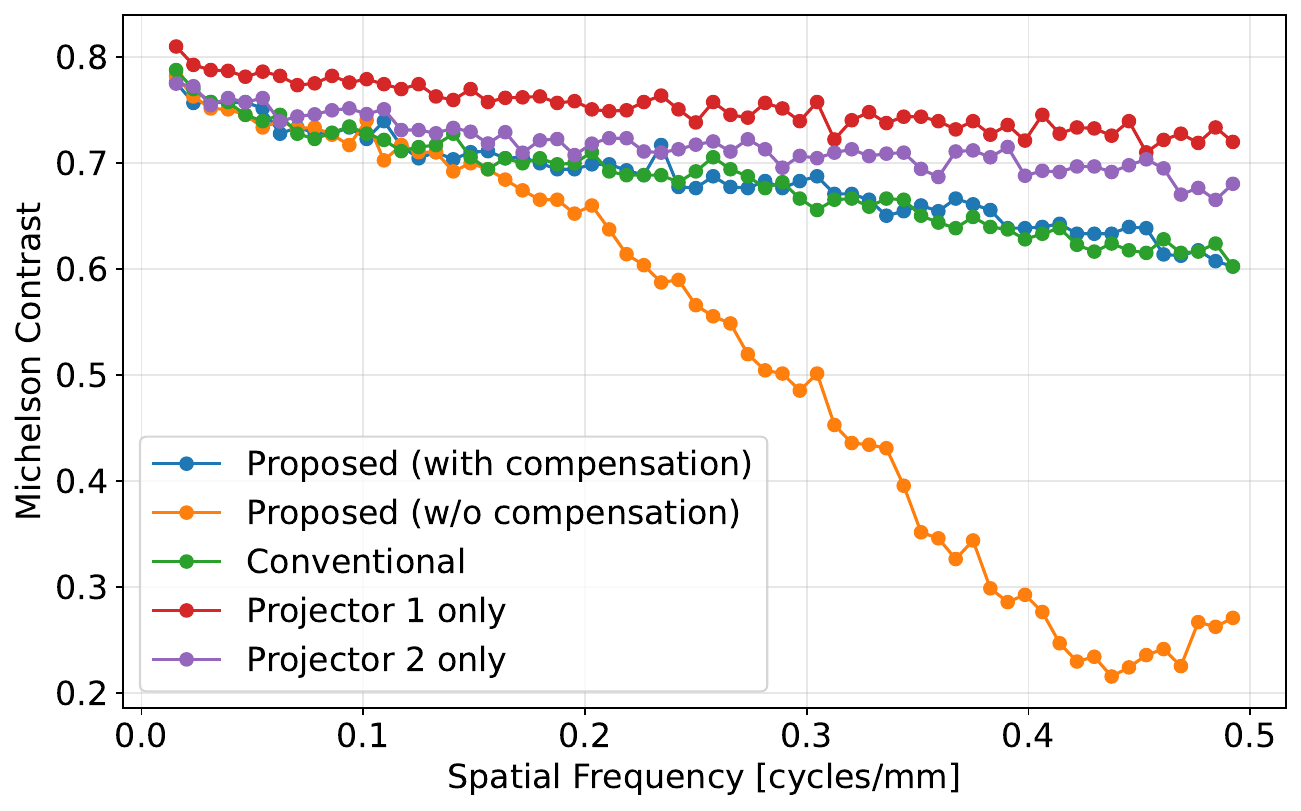}
    \caption{MTF comparison for the two-projector alignment.}
    \label{fig:mtf_comparison}
\end{figure}

To quantitatively evaluate alignment accuracy, we measured the modulation transfer function (MTF). 
Specifically, both projectors projected sinusoidal patterns whose brightness varied along the horizontal axis.
The spatial frequency was swept from $\frac{2}{128}$ cycles/mm to $\frac{63}{128}$ cycles/mm in increments of $\frac{1}{128}$ cycles/mm, and the contrast at each frequency was measured.
As a reference, the upper-limit MTF was measured by projecting the same pattern using only a single projector. 
The results (Fig.~\ref{fig:mtf_comparison}) show that without compensation, high-frequency components were significantly attenuated due to misalignment, whereas with compensation, the MTF closely matched the upper limit, demonstrating alignment accuracy equivalent to that of the conventional method.

\subsubsection{Twenty-Five Projectors}\label{subsubsec:25-pros}

\begin{figure}[tb]
    \centering
    \begin{subfigure}[t]{0.386\columnwidth}
        \centering
        \includegraphics[width=\textwidth]{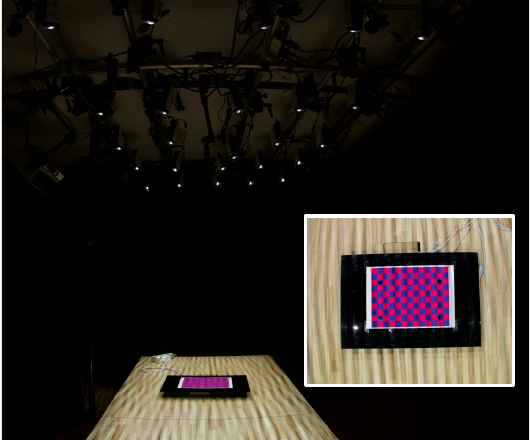}
        \caption{}
        \label{fig:25proj_checkerpattern_conventional}
    \end{subfigure}
    \hfill
    \begin{subfigure}[t]{0.573\columnwidth}
        \centering
        \includegraphics[width=\textwidth]{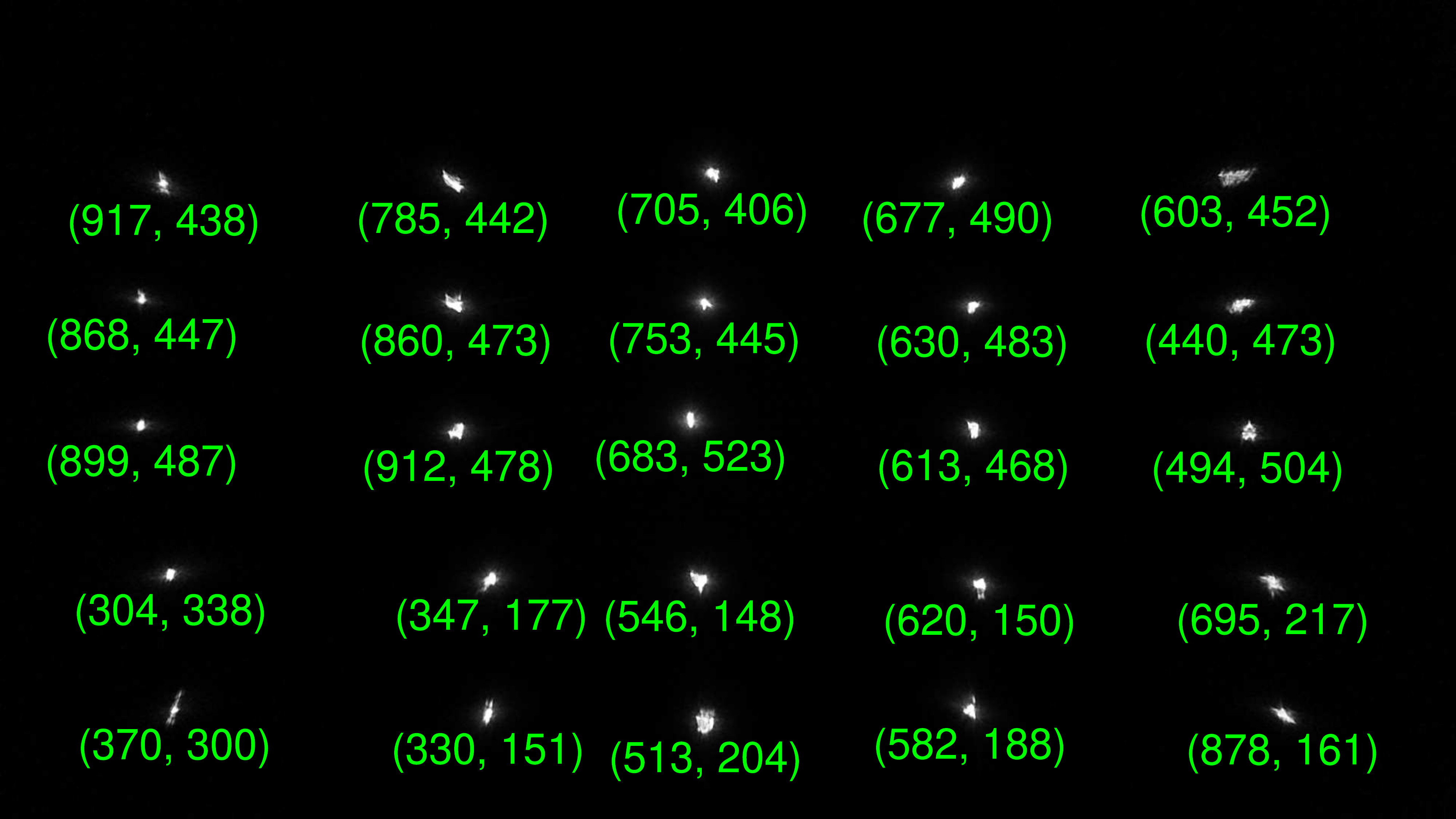}
        \caption{}
        \label{fig:25proj_checkerpattern_corrected}
    \end{subfigure}
    
    \vspace{0.5em}

    \begin{subfigure}[t]{0.32\columnwidth}
        \centering
        \includegraphics[width=\textwidth]{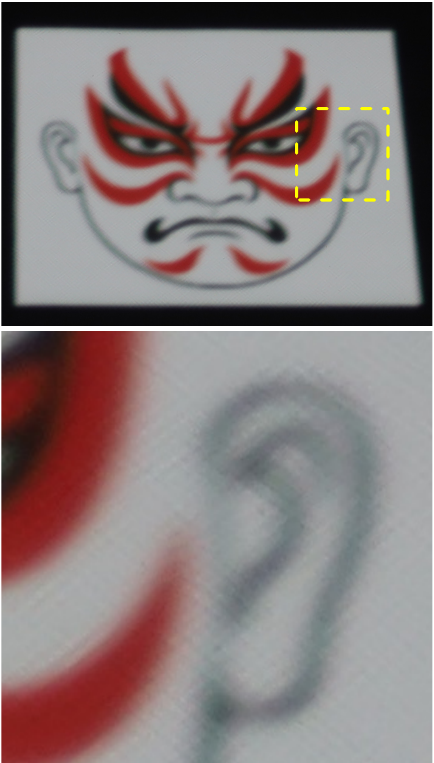}
        \caption*{\centering Conventional}
        \label{fig:25proj_texture_conventional}
    \end{subfigure}
    \hfill
    \begin{subfigure}[t]{0.32\columnwidth}
        \centering
        \includegraphics[width=\textwidth]{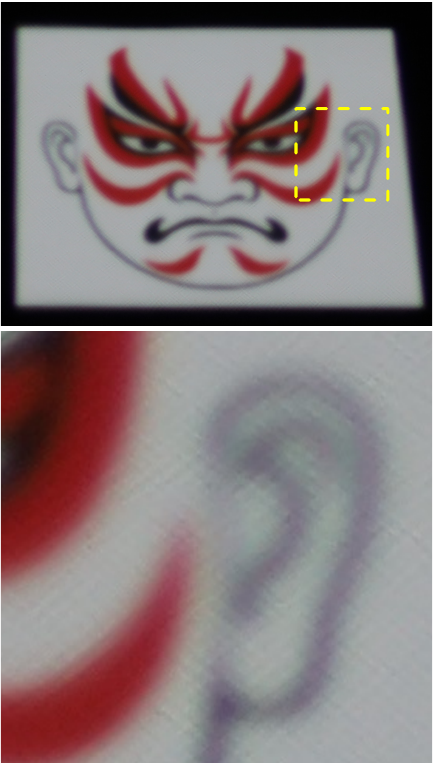}
        \caption*{\centering Proposed\\(w/o compensation)\\(c)}
        \label{fig:25proj_texture_uncorrected}
    \end{subfigure}
    \hfill
    \begin{subfigure}[t]{0.32\columnwidth}
        \centering
        \includegraphics[width=\textwidth]{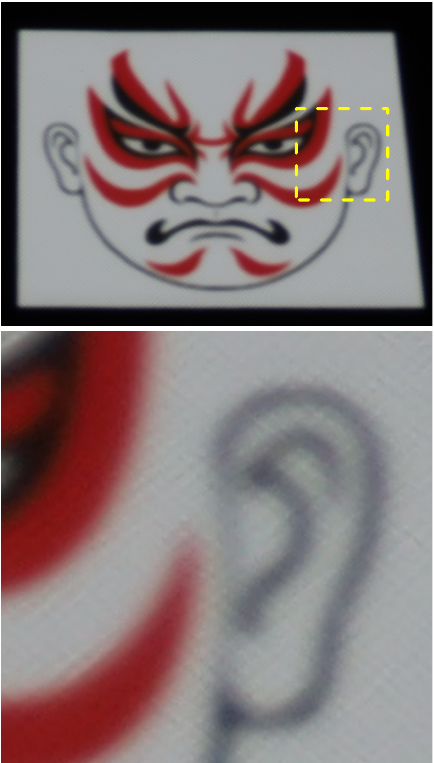}
        \caption*{\centering Proposed\\(with compensation)}
        \label{fig:25proj_texture_corrected}
    \end{subfigure}
    \caption{Experiment with 25 projectors. (a) The calibration board under simultaneous Gray-code projection. (b) Corresponding image captured by one of the embedded cameras with decoded coordinates overlaid. (c) Result of the alignment projection.}
    \label{fig:25proj_alignment}
\end{figure}

We further evaluated the scalability of the proposed method using a 25-projector PM system (Optoma ML1050ST+) arranged in a $5\times5$ ceiling-mounted array enabling shadowless PM~\cite{10.1145/3681756.3697948}. 
Gray-code and line-shift patterns were simultaneously projected from all projectors onto a calibration board placed on a table, and homographies were estimated both with and without the compensation for optical center misalignment, following the same procedure as in Section~\ref{subsubsec:2-pros}. 
Figure~\ref{fig:25proj_alignment} shows a simultaneous Gray-code projection from 25 projectors and the corresponding images captured by the embedded cameras. 
For comparison, we also estimated homographies under the \textit{conventional} condition by projecting structured light patterns from each projector individually and capturing them with an external camera (Basler a2A2590-60umBAS, 2592$\times$1944 pixels).

For evaluation, a white screen was attached to the board, and a facial illustration image---representing a typical PM application---was projected from all projectors using the computed homographies.
As shown in Fig.~\ref{fig:25proj_alignment}, the proposed method with compensation achieved a significantly sharper and clearer image than both the version without compensation and the conventional condition.
The difference between the compensated and uncompensated results demonstrates that the compensation effectively corrected the geometric misalignment caused by the offset of the embedded cameras' optical centers.
Furthermore, the fact that the compensated proposed method outperformed the conventional condition is likely because, in the conventional setup, the external camera was mounted near the ceiling alongside the projectors, resulting in a long distance to the calibration board. This reduced the apparent spatial resolution and degraded alignment accuracy. In contrast, the proposed method relies on cameras embedded directly in the board, avoiding this loss of apparent resolution and thus enabling higher-precision alignment.

\subsubsection{Summary}

The results from the experiments in Sections~\ref{subsubsec:2-pros} and~\ref{subsubsec:25-pros} confirm that the proposed method achieves projection alignment accuracy comparable to or even better than the conventional approach when the compensation for optical center misalignment is applied, demonstrating the effectiveness of the compensation technique.
In the 25-projector experiment, the conventional approach required sequential structured light projection from all projectors, involving 44 patterns per projector (1,100 patterns total) and taking approximately 12 minutes.
The proposed method, on the other hand, required only 54 projected patterns through simultaneous projection from all devices, resulting in a 95\% reduction in pattern count.

\subsection{Projector Calibration}
\label{subsec:pro-calib}
We next evaluated the proposed method's ability to estimate both the intrinsic and extrinsic parameters of multiple projectors. 
We calibrated three projectors with different models (Optoma ML1050ST+, BenQ TK685, and BenQ TK850). 
The calibration board was positioned in 8 different poses, and at each pose, all projectors simultaneously projected structured light patterns, which were captured by the embedded cameras to obtain correspondences between projector and board coordinates. 
Using Zhang's algorithm~\cite{888718}, the intrinsic and extrinsic parameters of each projector were then estimated under two conditions: with and without the compensation for optical center misalignment. 
For comparison, projector calibration was also performed under the \textit{conventional} condition using the external camera and the same board poses. 
Under the conventional condition, we conducted calibration using two settings: one using all 9~$\times$~12 checkerboard corners (108-corner condition), and another using only four checkerboard corners, matching the corner count available in the proposed method (4-corner condition).

\begin{table}[tb]
    \centering
    \caption{RMS reprojection errors (pixels) for each projector}
    \label{tab:reprojection_errors}
    \resizebox{\columnwidth}{!}{%
    \begin{tabular}{lcccc}
        \hline
        & Conventional
        & Conventional        
        & Proposed
        & Proposed \\ 
        Projector 
        & (108-corner)
        & (4-corner)        
        & (w/o compensation)
        & (with compensation) \\ 
        \hline
        ML1050ST+ & 0.34 & 0.65 & 2.18 & 0.91 \\
        TK685 & 0.38 & 0.76 & 2.47 & 0.91 \\
        TK850 & 0.39 & 0.75 & 2.44 & 0.89 \\
        \hline
    \end{tabular}%
    }
\end{table}

\begin{figure}[tb]
    \centering
    \includegraphics[width=\linewidth]{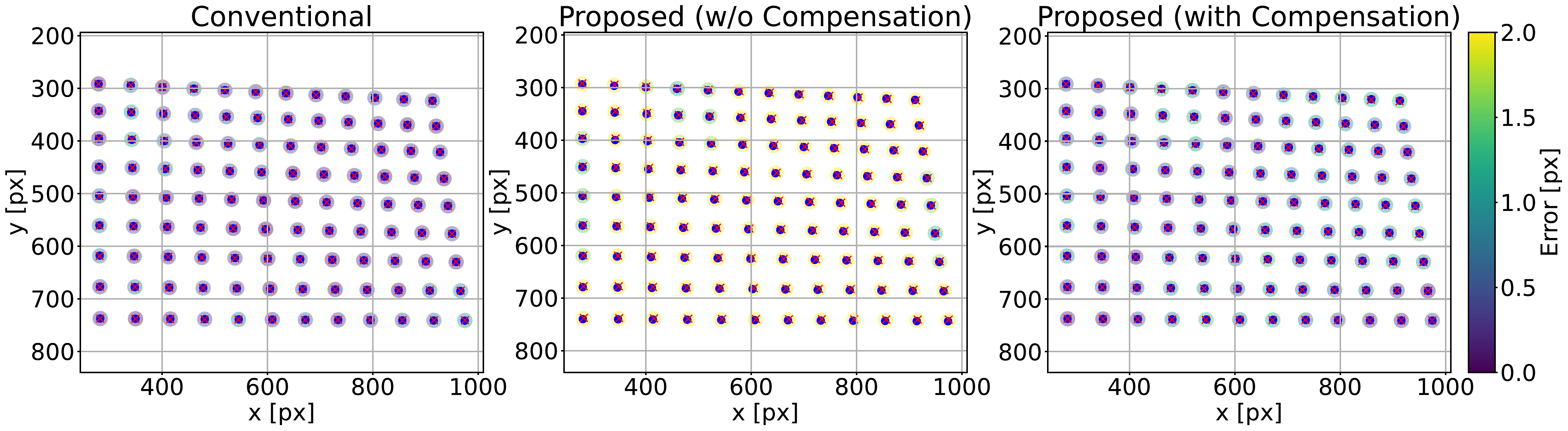}
    \caption{Reprojection Errors at checker corners. (\color{blue}{\textbullet}\color{black}: Projected, \color{red}{$\times$}\color{black}: Measured)}
    \label{fig:reprojection_error_comparison}
\end{figure}



Table~\ref{tab:reprojection_errors} summarizes the reprojection errors.
Figure~\ref{fig:reprojection_error_comparison} shows the error at each checker corner for a certain board pose.
The reprojection errors were below 1.0~pixel in both the proposed method with compensation for optical center misalignment and the conventional method, which is generally regarded as sufficiently accurate for geometric calibration. 
In contrast, without compensation, the reprojection error increased to around 2.4~pixels in the proposed method, demonstrating the necessity of the compensation step for achieving high accuracy.
The difference in accuracy between the proposed and conventional methods suggests that calibration accuracy could be further improved by increasing the number of embedded cameras, as evidenced by the higher accuracy under the 108-corner condition.
While both methods require multiple board relocations, the conventional approach requires sequential pattern projection for every projector, whereas the proposed method does not, resulting in a substantial reduction in time and effort.

\subsection{Robustness to Ambient Light}
\label{subsec:ambient-light}

\begin{figure}[tb]
    \centering
   \includegraphics[width=\linewidth]{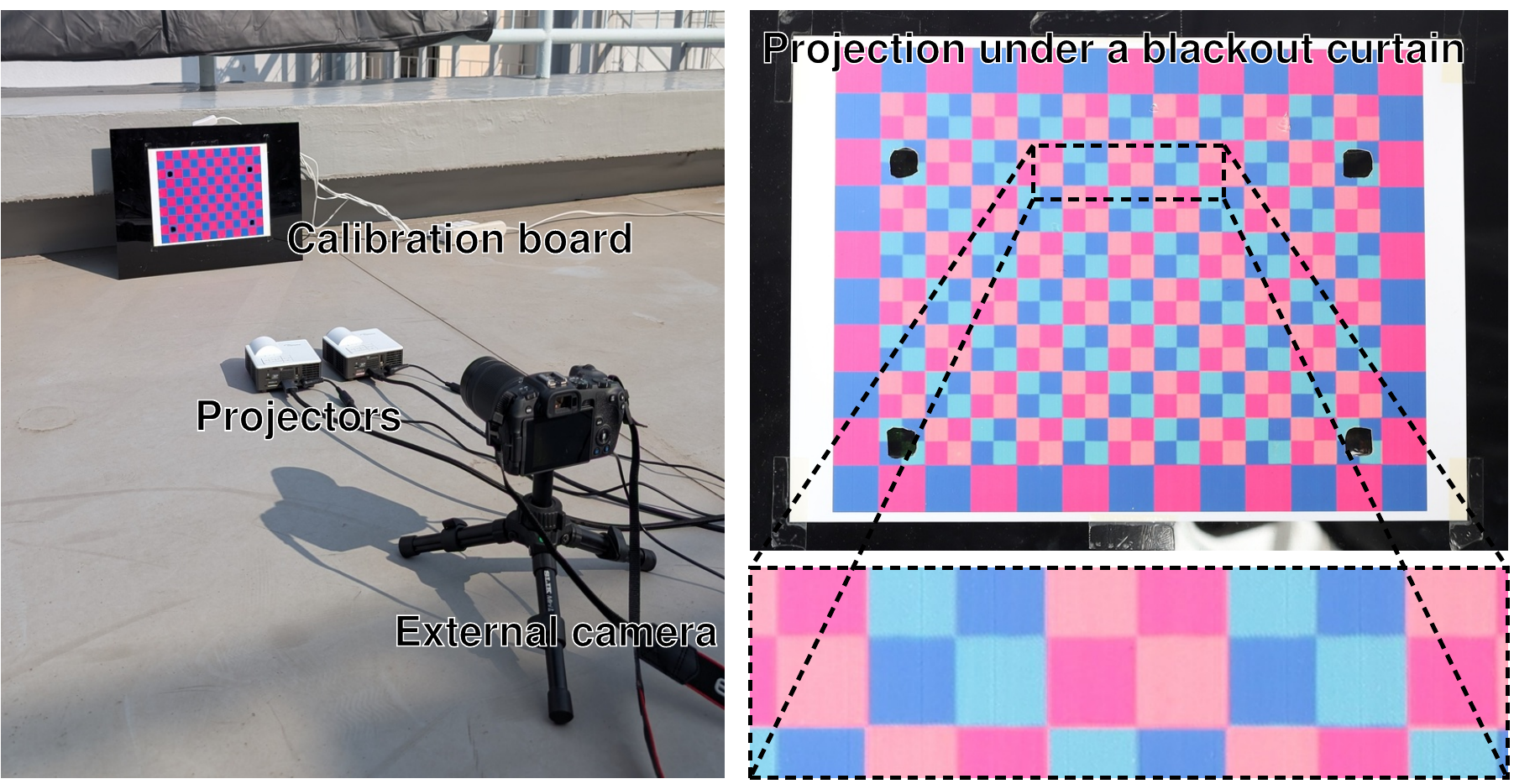}
    \caption{Experiment evaluating robustness to ambient light with two projectors outdoors. (Left) Experimental setup placed outdoors under direct sunlight ($\approx$70\,klux). (Right) Alignment result of a half-scale checkerboard pattern using the homography transformation estimated with the proposed method.}
    \label{fig:outdoor_experiment}
\end{figure}

Finally, we evaluated robustness to strong ambient illumination with two simultaneously active projectors (Optoma ML1050ST+) set up outdoors under approximately 70\,klux of sunlight (Fig.~\ref{fig:outdoor_experiment} left), as measured with a MINOLTA T-10 illuminance meter.
Gray-code and line-shift patterns were simultaneously projected from both projectors; the per-projector patterns were successfully separated and decoded, and the compensation for optical center misalignment was applied to estimate each homography.
For comparison, homographies were also estimated using the conventional external camera-based method.

For evaluation, the setup was covered with a blackout curtain and a half-scale checkerboard was projected from each projector to assess positional alignment.
The proposed method achieved accurate alignment (Fig.~\ref{fig:outdoor_experiment} right), whereas the conventional method failed entirely: the projected intensity variations were overwhelmed by the ambient light, preventing structured light pattern detection.
These results confirm that the proposed method can robustly separate and decode patterns from multiple simultaneously active projectors even under extreme outdoor illumination.

\section{Conclusion}

In this study, we tackled the scalability limitation in projector calibration by embedding cameras into the calibration target.
This configuration enables the separation and measurement of structured light patterns simultaneously projected from multiple projectors according to their incident directions.
We also introduced a compensation method to correct errors caused by the optical centers of the embedded cameras not being perfectly aligned with the calibration surface.
Experimental results using a prototype system demonstrated that the proposed method significantly reduces the number of required projection patterns while achieving calibration accuracy comparable to conventional approaches using external cameras.

The embedded cameras capture not only positional but also directional information, analogous to plenoptic cameras~\cite{ng2005light}.
In the future, we plan to leverage this directional data as an additional constraint in parameter estimation to further reduce the number of required calibration board poses.
We also plan to investigate combining our method with single-shot structured light approaches (e.g., De Bruijn sequences) to reduce the per-projector pattern count, and to extract further physical properties such as lens aberration characteristics, aperture shape, and focusing distance.

\paragraph{Acknowledgements.}
This work was supported by JSPS KAKENHI Grant Number JP25K03155 and JST ASPIRE Grant Number JPMJAP2404.
{
    \small
    \bibliographystyle{ieeenat_fullname}
    \bibliography{main}
}


\end{document}